# Matching of Images with Rotation Transformation Based on the Virtual Electromagnetic Interaction

XIAODONG ZHUANG[1,2] and N. E. MASTORAKIS[1]

1. Technical University of Sofia, Industrial Engineering Department, Kliment Ohridski 8, Sofia, 1000 BULGARIA
   (mastor@wseas.org, http://www.wseas.org/mastorakis)
2. Qingdao University, Automation Engineering College, Qingdao, 266071 CHINA
   (xzhuang@worldses.org, http://research-xzh.cwsurf.de/)

*Abstract:* -A novel approach of image matching for rotating transformation is presented and studied. The approach is inspired by electromagnetic interaction force between physical currents. The virtual current in images is proposed based on the significant edge lines extracted as the fundamental structural feature of images. The virtual electromagnetic force and the corresponding moment is studied between two images after the extraction of the virtual currents in the images. Then image matching for rotating transformation is implemented by exploiting the interaction between the virtual currents in the two images to be matched. The experimental results prove the effectiveness of the novel idea, which indicates the promising application of the proposed method in image registration.

## 1 Introduction

For a long time, researchers have been obtaining inspirations from nature for designing new methods to solve difficult problems. In computer science, such nature-inspired methods have achieved notable success, such as Genetic Algorithm, Ant Colony Optimization, Artificial Neural Network, etc. In image processing, the nature-inspired methodology has also been applied. Such methods are somewhat like virtual experiments by modeling and simulating the natural mechanisms upon virtual objects (such as the digital image) on computer platform, and the experimental results are studied and exploited in solving problems.

Physics inspired methods have become a novel branch in image processing in recent years. Electro-magnetic field inspired method is one important category of such methods, and promising results have been obtained in edge detection, corner detection, shape skeletonization, ear recognition, etc [1-8]. Most methods in previous research are mainly inspired by electro-static field, but the analysis and imitation of magnetic field is much less. Moreover, almost all previous research concentrated on the imitation of static fields, but the dynamic interaction between images was not investigated to the authors' knowledge. In this paper, the virtual electro-magnetic interaction in digital images is studied and applied in image matching for rotating transformation.

Image matching is an important practical task in image processing [9-17]. Based on the matching result of two related image, the integration of useful data can be achieved in the two related but separated images. Currently, some matching methods are based on grayscale or color, while others are based on feature extraction [9-17]. The nature of transformation from one image to another may be one of the following cases: rigid, affine, projective or curved [9-17]. Different methods have their own advantages and disadvantages respectively. Currently, it is relatively difficult to implement the matching both very rapidly and very accurately (on commonly-used PCs). Moreover, most methods are not full-automatic because they need manual intervention. Traditional matching methods are based on the similarity between the matched areas in two images. Quite different from previous matching methods, in this paper a novel idea for matching is presented based on the virtual interaction force which can reflect the transformation between two images. Inspired by the physical phenomena of the attraction between two current-carrying wires, an automatic image matching method is proposed for rotating transformation. In the method, first the significant edge lines are extracted as the structural feature of the images. Then one image produces a virtual magnetic field by the virtual edge current, and puts virtual magnetic force on each edge element the other image. The total moment is calculated by the summation of all the moments generated by the virtual forces respectively, which can represent the rotating transformation between the two images. Then the matching can be implemented by following the guidance of the total moment, which provides an efficient and convenient way of image matching for rotating transformation.

In the authors' earlier work, the force between the virtual currents in digital image has been studied and applied in matching for shifting transformation. Since rotation is a different image transformation from shifting, in this paper the virtual moment produced by the force between virtual currents is further studied, based on which a method of image matching for rotation transformation is proposed. The experimental results indicate that the virtual moment method is just suitable to the matching for rotation transformation, which is an extension of the previous virtual current method to solve that new problem. Moreover, in the experiments, its effectiveness for the case of large rotation angle is distinctive, which indicates its potential value in applications.

## 2 The Electro-Magnetic Interaction between Physical Currents

### 2.1 The magnetic field of physical current

In physics, the description of the magneto-static field is given by the Biot-Savart law [18-21], where the source of the magnetic field is the current of arbitrary shapes which is composed of current elements. A current element $I\vec{dl}$ is a vector representing a very small part of the whole current, whose magnitude is the arithmetic product of $I$ and $dl$ (the length of a small section of the wire). The current element has the same direction as the current flow on the wire. The magnetic field generated by a current element $I\vec{dl}$ is as following [18-21]:

$$\vec{dB} = \frac{\mu_0}{4\pi} \cdot \frac{I\vec{dl} \times \vec{r}}{r^3} \quad (1)$$

where $\vec{dB}$ is the magnetic induction vector at a space point. $I\vec{dl}$ is the current element. $r$ is the distance between the space point and the current element. $\vec{r}$ is the vector from the current element to the space point. The operator $\times$ represents the cross product. The direction of the magnetic field follows the right-hand rule [18-21]. The direction distribution of the magnetic field on the 2D plane where the current element lies is shown in Fig. 1, together with a 3D demonstration.

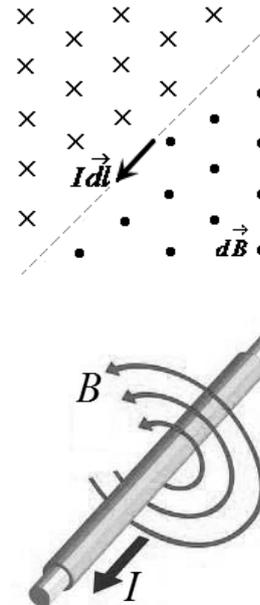

**Fig. 1 The magnetic field's direction distribution of a current element**

The magnetic field generated by the current in a wire of arbitrary shape is the accumulation of the fields generated by all the current elements on the wire, which is described by the Biot-Savart law [18-21]:

$$\vec{B} = \int_D \vec{dB} = \int_D \frac{\mu_0}{4\pi} \cdot \frac{I\vec{dl} \times \vec{r}}{r^3} \quad (2)$$

where $\vec{B}$ is the magnetic induction vector at a space point generated by the whole current of the wire. $D$ is the area where the current element exists. $\vec{dB}$ is the magnetic field generated by each current element in $D$.

### 2.2 The force on physical current in stable magnetic field

In electro-magnetic theory, the magnetic field applies force on moving charges, which is described by the Lorentz force. Derived from the Lorentz force, the magnetic force on a current element is as following [18-21]:

$$d\vec{F} = I\vec{dl} \times \vec{B} \quad (3)$$

where $Idl$ is the current element and $dF$ is the magnetic force on it caused by the magnetic induction *B*. The direction of *dF* satisfies the "left hand rule".

A current-carrying wire of arbitrary shape consists of many current elements. The magnetic force on a wire is the summation of the force *dF* on all its current elements, which is as following [18-21]:

$$\vec{F} = \int_C I d\vec{l} \times \vec{B} \qquad (4)$$

where *C* is the integration path along the wire, and *F* is the total magnetic force on the wire.

## 2.3 The electromagnetic interaction between two physical currents

The current-carrying wire can generate magnetic field, meanwhile the magnetic field can put force on another wire. Thus there is interaction force between two current-carrying wires. In electro-magnetic theory, the Ampere force between two wires of arbitrary shapes is based on the line integration and combines the Biot-Savart law and Lorentz force in one equation as following :

$$\vec{F}_{12} = \frac{\mu_0}{4\pi} \int_{C_1} \int_{C_2} \frac{I_1 d\vec{l}_1 \times (I_2 d\vec{l}_2 \times \vec{r}_{21})}{r_{21}^3} \qquad (5)$$

where $F_{12}$ is the total force on wire1 due to wire2. $\mu_0$ is the magnetic constant. $I_1 dl_1$ and $I_2 dl_2$ are the current elements on wire1 and wire 2 respectively. $r_{21}$ is the vector from $I_2 dl_2$ to $I_1 dl_1$. $C_1$ and $C_2$ are the integration path along the two wires respectively.

Many physical experiments of electromagnetic interaction have obtained interesting results. For example, consider the following experiment shown in Fig. 2. There are two line wires $I_1$ and $I_2$. The position of $I_2$ is fixed and it can not move. $I_1$ hangs above $I_2$, and $I_1$ can rotate around the point $O'_1$. If there are currents in the two straight wires, there will be force and also moment on $I_1$ so that $I_1$ will rotate to the position at which the angle between the directions of the two currents becomes zero (i.e. their directions become the same). This experiment indicates the location-restoring effect for two wires with rotation transformation, which may be exploited in image matching.

Equation (5) is virtually the accumulation of all the magnetic forces on the current elements in wire1 put by the current elements in wire2. If such interaction is simulated on computers, the continuous wires should be discretized, and the integration in Equation (5) should be discretized to summation:

$$\vec{F}_d = \frac{\mu_0}{4\pi} \sum_{\vec{T}_{1j} \in C_1} \sum_{\vec{T}_{2k} \in C_2} \frac{\vec{T}_{1j} \times (\vec{T}_{2k} \times \vec{r}_{kj})}{r_{kj}^3} \qquad (6)$$

where $F_d$ is the force on wire1 from wire2. Here both wire1 and wire2 are in discrete form, which consist of a set of discrete current element vectors respectively. $C_1$ and $C_2$ are the two sets of discrete current elements for wire1 and wire2 respectively. In another word, all the discrete vectors in $C_1$ constitute the discrete form of wire1, and all the discrete vectors in $C_2$ constitute the discrete form of wire2. $T_{1j}$ and $T_{2k}$ are the current element vectors in $C_1$ and $C_2$ respectively. $r_{kj}$ is the vector from $T_{2k}$ to $T_{1j}$.

Some examples of discretized current-carrying wires are shown in Fig. 3. In computer simulations, the continuous current directions are discretized into 8 directions: {east, west, north, south, northeast, northwest, southeast, southwest}. Fig. 3(a) shows the discrete form of a continuous current with the rectangle shape. Fig. 3(b) shows the discrete form of a continuous current with the circle shape. Fig. 3(c) shows the discrete form of a continuous current with the straight-line shape.

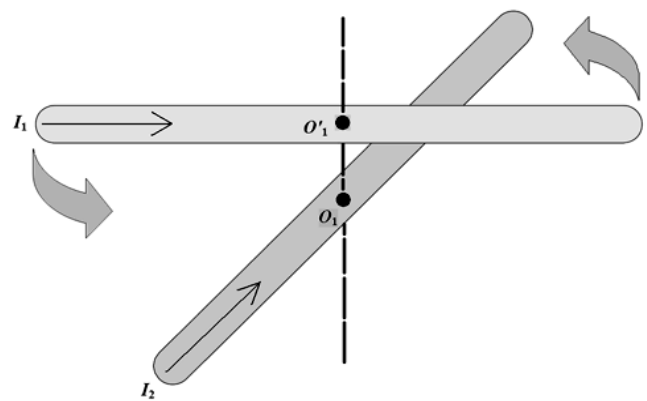

**Fig. 2　An example experiment of electromagnetic interaction**

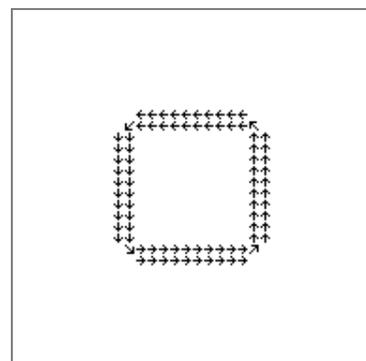

**Fig. 3(a) The discrete form of a continuous rectangle current**

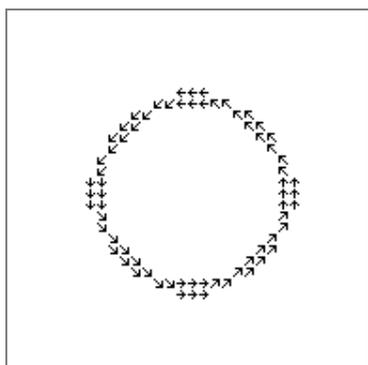

**Fig. 3(b) The discrete form of a continuous circle current**

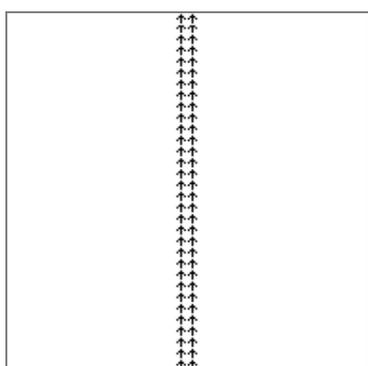

**Fig. 3(c) The discrete form of a continuous straight-line current**

## 3 The Virtual Currents and Their Interaction in Digital Images

Among the current methods of image matching, feature-based methods have the notable advantage of much less computation load. There are several important image features which can represent image structure, and edge is one of them. On the other hand, many physical experiments (such as Fig. 2) and theoretical analysis indicate that the electro-magnetic interaction can cause a rotated current-carrying wire attracted to another wire of the same shape on the original location. (Strictly speaking, the attraction between the fixed wire and the moved one exists only in a certain range of position deviation). If a proper line feature representing the image structure can be found in the image, it is possible to exploit the above location-restoring effect of electro-magnetic interaction in image matching. The edge feature is just suitable for this requirement.

### 3.1 Extraction of the "significant edge current"

In this paper, the "significant edge points" are extracted for image matching. The significant edges are definite borders of regions, and there is sharp change of grayscale across the significant edge lines. In the proposed matching method, the significant edge points are first extracted in the two images to be matched. The virtual currents in the image are defined as the set of discrete current elements on the significant edge lines. Then the interaction force between the virtual currents in the two images is calculated, which inspires a novel matching approach for rotating transformation.

Canny operator is widely used to extract edge lines [22,23]. To improve computation efficiency, in this paper a simplified Canny-like method is proposed to extract the significant edges in image. First, Sobel operator is used to estimate the gradient fields of the two images to be matched. Then the threshold processing of the gradient magnitude is performed to reserve the definite edge points, in which the points with a magnitude smaller than the threshold value are eliminated from the set of edge points. (The threshold value is set as a predefined percent of the maximum value of gradient magnitude.) After that, a "non-maximum suppression" is performed to get thin edge lines. For an edge point, it is reserved in the set of significant edge points only if its gradient magnitude is larger than the adjacent points on at least two of the following pairs of directions: west and east, north and south, northwest and southeast, northeast and southwest.

With the above process, the significant edge lines can be extracted. However, the direction of current element should be along the tangent direction of the wire, while the gradient vector is perpendicular to the tangent direction of the edge line. To get the virtual edge current, all the gradient vectors on the significant edge lines rotate 90 degrees. Then the vector direction after rotation is along the tangent direction of edge lines. In another word, on the significant edge lines, the virtual current elements are obtained by rotating the gradient vectors 90 degrees counterclockwise. A simple example is shown in Fig. 4.

All the virtual current element vectors on the significant edges form the virtual currents in the image. In another word, the virtual edge current in an image is a set of discrete current element vectors on the significant edge lines, whose amplitudes (or amperage) are equal to their corresponding gradients respectively, while their directions are just perpendicular to their corresponding gradient vectors respectively.

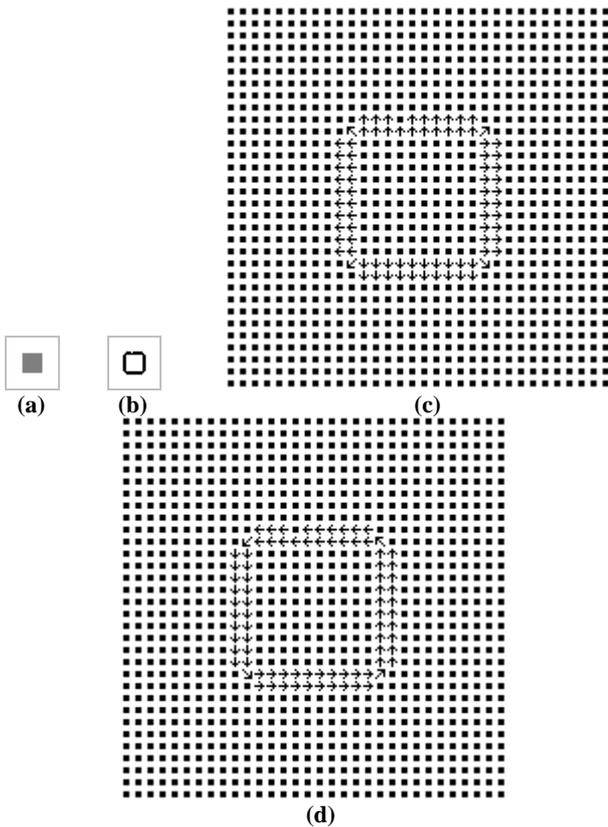

Fig. 4 An example of the significant edge lines and the corresponding virtual current of a square shape
(a) The image of the square shape
(b) The significant edge lines of (a)
(c) The gradient vectors on the significant edge lines
(d) The discrete virtual current

Fig. 4(a) shows a simple image of a square. Fig. 4(b) shows its significant edge lines extracted. Fig. 4(c) shows the direction distribution of the discrete gradient vectors, where the continuous gradient direction is discretized into eight directions: {east, west, north, south, northeast, northwest, southeast, southwest}. Fig. 4(d) shows the direction distribution of the discrete current elements, which is the rotated version of gradient vector. The dots in Fig. 4(c) and Fig. 4(d) show the points with no current elements.

Some examples of extracting virtual edge current are shown from Fig. 5 to Fig. 9. Fig. 5(a) to Fig. 8(a) show some simple images of the size $32 \times 32$. Some real world images of the size $128 \times 128$ are shown in Fig. 9. The significant edge lines extracted are also shown in these results, which will be used as virtual edge currents in the following matching process.

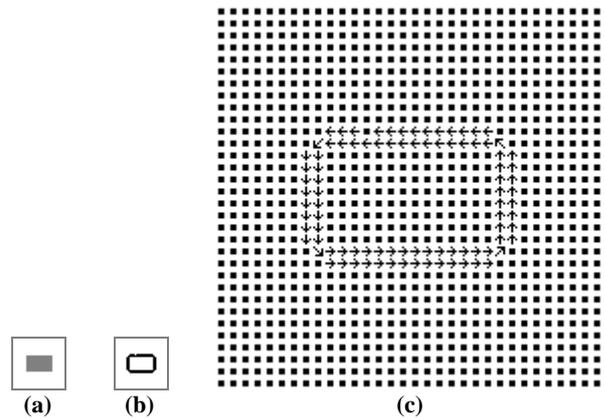

Fig. 5 The significant edge lines and the corresponding virtual current of a rectangle shape
(a) The image of the rectangle shape
(b) The significant edge lines of (a)
(c) The discrete virtual current

Fig. 5(a) shows an image of a rectangle. Fig. 5(b) shows its significant edge lines. Fig. 5 (c) shows the virtual current elements on the significant edge lines, where the arrows show the discrete directions of discrete current elements. The dots in Fig. 5(c) show the points with no current elements.

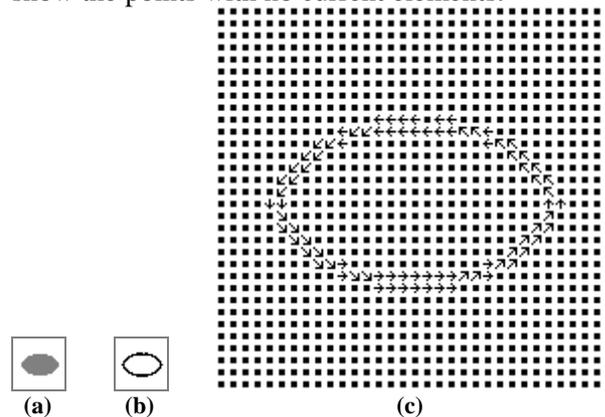

Fig. 6 The significant edge lines and the corresponding virtual current of an ellipse shape
(a) The image of the ellipse shape
(b) The significant edge lines of (a)
(c) The discrete virtual current

Fig. 6(a) shows an image of an ellipse. Fig. 6(b) shows its significant edge lines. Fig. 6(c) shows the virtual current elements on the significant edge lines, where the arrows show the discrete directions of discrete current elements.

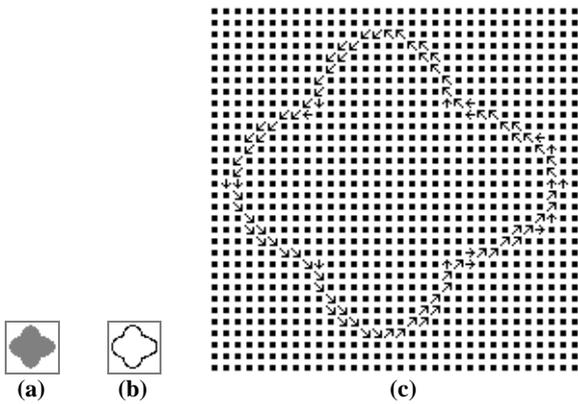

**(a)　　　(b)　　　　　　　　　　(c)**
**Fig. 7 The significant edge lines and the corresponding virtual current of an irregular shape**
**(a) The image of the irregular shape**
**(b) The significant edge lines of (a)**
**(c) The discrete virtual current**

　　Fig. 7(a) shows an image of an irregular shape. Fig. 7(b) shows its significant edge lines. Fig. 7(c) shows the virtual current elements on the significant edge lines, where the arrows show the discrete directions of discrete current elements.

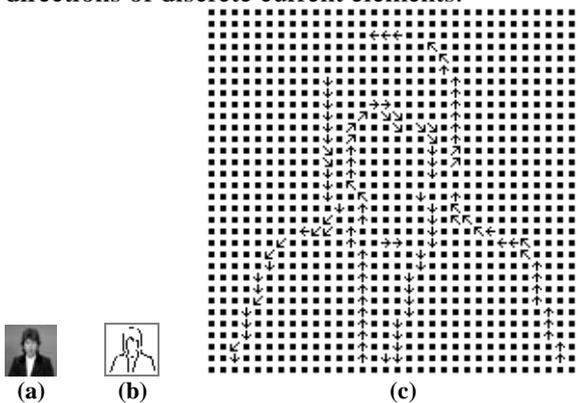

**(a)　　　(b)　　　　　　　　　　(c)**
**Fig. 8 The significant edge lines and the corresponding virtual current of a shrunken image of broadcaster**
**(a) The shrunken broadcaster image**
**(b) The significant edge lines of (a)**
**(c) The discrete virtual current**

　　Fig. 8(a) shows the shrunken version of a broadcaster image. Fig. 8(b) shows its significant edge lines. Fig. 8(c) shows the virtual current elements on the significant edge lines. In Fig. 9, the real world images and their significant edge lines are shown, including the images of the broadcaster, the cameraman, the peppers, the locomotive, the boat and a medical image of brain.
　　The virtual current elements are defined according to the gradient in the image, the consistency (or matching) of the current amperage is guaranteed by the gradient along each "significant edge line" extracted in the proposed method. Since the gradient estimated is for digital images, the virtual current elements are just discretized form on the 2D image plane.

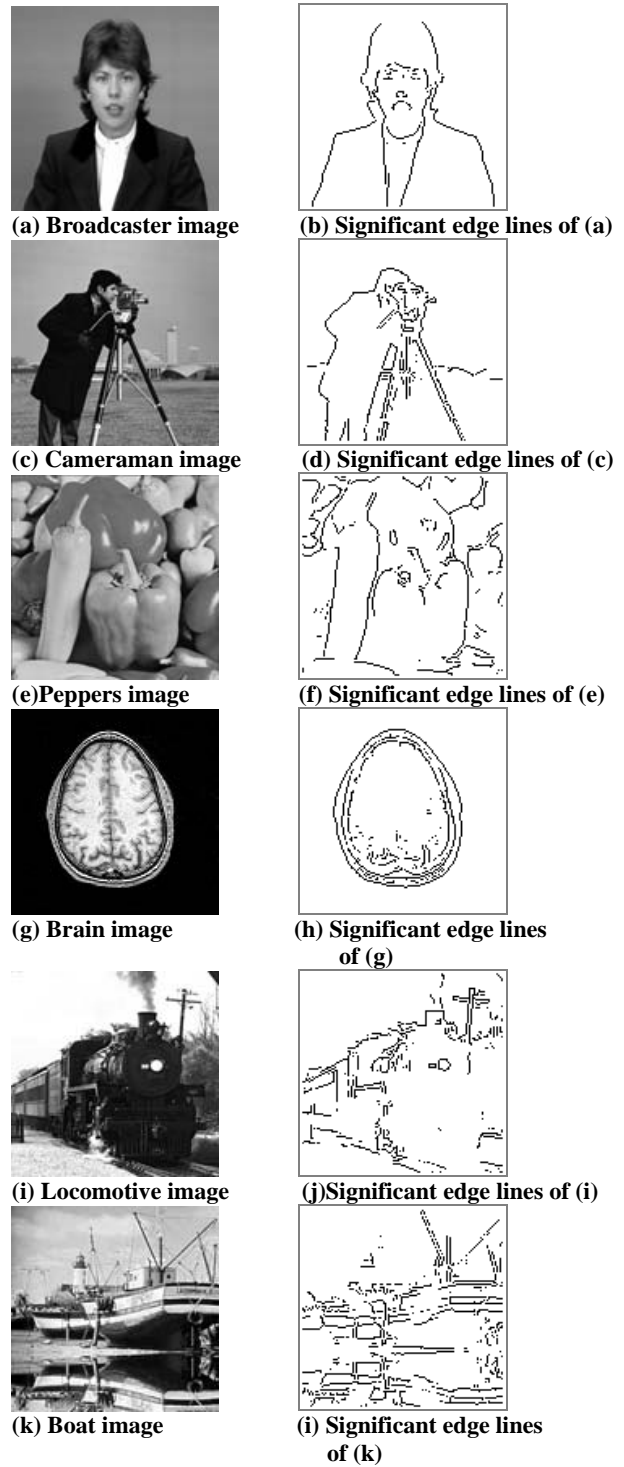

**(a) Broadcaster image　　(b) Significant edge lines of (a)**
**(c) Cameraman image　　(d) Significant edge lines of (c)**
**(e) Peppers image　　(f) Significant edge lines of (e)**
**(g) Brain image　　(h) Significant edge lines of (g)**
**(i) Locomotive image　　(j) Significant edge lines of (i)**
**(k) Boat image　　(l) Significant edge lines of (k)**

**Fig. 9 A group of real world images and their significant edge lines**

## 3.2 The interaction between the virtual current elements in images

If the virtual currents in the two images are extracted respectively, the total interaction force between them can be calculated according to Equation (6). Suppose the sets of discrete current elements in image1 and image2 are $C_1$ and $C_2$ respectively. Each discrete current element $T_{1j}$ in $C_1$ is applied the force by all the current elements in $C_2$:

$$\vec{F}_{1j} = A \cdot \sum_{\vec{T}_{2k} \in C_2} \frac{\vec{T}_{1j} \times (\vec{T}_{2k} \times \vec{r}_{kj})}{r_{kj}^3} \quad (7)$$

where $F_{1j}$ is the force on $T_{1j}$ from $C_2$. $T_{2k}$ is a current element vector in $C_2$. $r_{kj}$ is the vector from $T_{2k}$ to $T_{1j}$. $A$ is a constant value. A simulation example of the force on the virtual current elements are shown in Fig. 10.

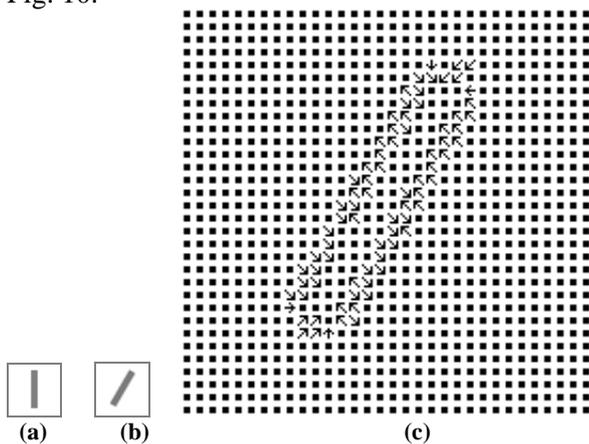

**Fig. 10 The forces on the current elements in the rotated rectangle image (30 degrees)**
(a) the original rectangle image
(b) the rotated image of (a) by 30 degrees
(c) the forces on the virtual current elements in the rotated image

Fig. 10(a) is the original image of the size 32×32. Fig. 10(b) shows the rotated image with 30 rotating degrees. Fig. 10(c) shows the force distribution in the rotated image. In Fig. 10(c), each arrow shows the discrete direction of the virtual force on the virtual current element at that position, and the black dots indicate the positions without virtual current element. If carefully observed, it can be estimated from Fig. 10(c) that most of the forces have the effect of rotating the rectangle in (b) counterclockwise to the original position shown in (a).

## 4 The Magnetic Moment on the Virtual Current in the Rotated Image

In physics, a force on an object can cause change in its motion by both shifting and rotating. The effect of rotation by a force is described by the moment. The moment of a force $F$ on an object with respect to the point $O$ is defined as following [24,25]:

$$\vec{M}_O = \vec{r}_{OF} \times \vec{F} \quad (8)$$

where $r_{OF}$ is the vector from point $O$ to the position where the force $F$ is applied, and × represents the cross product. In the right-handed coordinate system, the direction of the moment and the rotating direction caused by it obey the right hand rule shown as Fig. 11. If the positive direction of the moment is defined as coming out of the paper, the positive moment will cause the counter-clockwise rotation.

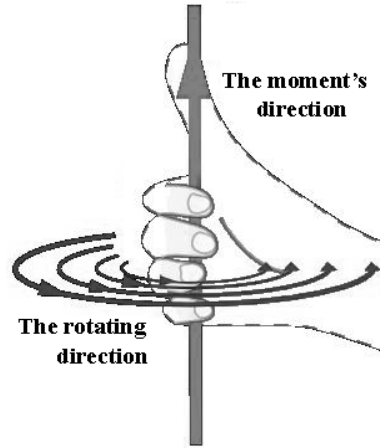

**Fig. 11 The right hand rule**

But in the left-handed coordinate system such as the screen coordinate used here, it is different that the negative moment computed by Equation (8) under the left-handed coordinate will cause the counter-clockwise rotation. The screen coordinate is shown in Fig. 12.

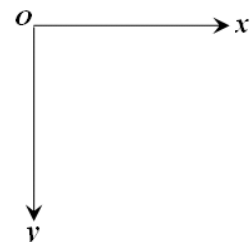

**Fig. 12 The screen coordinates**

It is interesting to study the effect of the moment between two images with rotating transformation. In the interaction between two current-carrying wires, each current element in one wire has force applied on it from the other wire. The force on each current element also generates a moment, which can cause the rotation of the wire. Physical experiments such as Fig. 2 inspire the idea that it is worthwhile to study the interaction between two wires to design novel image matching approach for rotating transformation, where the interaction has the effect of rotating $I_1$ to the same direction of $I_2$.

The force on the current elements in one image by the other is proposed in Equation (7). The force on each discrete current element also generate the moment as following (here suppose the two images are on the same plane):

$$\vec{M}_{Oj} = \vec{r}_{OF_{1j}} \times \vec{F}_{1j} = \vec{r}_{OF_{1j}} \times \sum_{\vec{T}_{2k} \in C_2} \frac{\vec{T}_{1j} \times (\vec{T}_{2k} \times \vec{r}_{kj})}{r_{kj}^3}$$

(9)

where $M_{Oj}$ is the moment on a current element $T_{1j}$ caused by the virtual current $C_2$. $C_2$ is the virtual current in image2. In this paper, only the rotation around the image center is investigated. Thus $O$ is the center point of image1. $F_{1j}$ is the force on $T_{1j}$ caused by the virtual current $C_2$. $r_{OF1j}$ is the vector from point $O$ to the position where the force $F_{1j}$ is applied.

The total moment on image1 is the accumulation of all the moments on the current elements in image1:

$$\vec{M}_{O1} = \sum_{\vec{T}_{1j} \in C_1} \vec{r}_{OF_{1j}} \times \vec{F}_{1j} = \sum_{\vec{T}_{1j} \in C_1} \vec{r}_{OF_{1j}} \times \sum_{\vec{T}_{2k} \in C_2} \frac{\vec{T}_{1j} \times (\vec{T}_{2k} \times \vec{r}_{kj})}{r_{kj}^3}$$

(10)

where $M_{O1}$ is the total moment on the virtual current $C_1$ caused by the virtual current $C_2$. $C_1$ and $C_2$ are the virtual currents in image1 and image2 respectively.

The experiments have been carried out for test images as well as real world images. An example of the experiments is shown as following. An image of a rectangle is rotated clockwise around the image center by 30 degrees and 45 degrees respectively. The virtual force on each virtual current element in the rotated image is calculated. Then the total moment is calculated according to Equation (3). The experimental results are shown in Fig. 13 and Fig. 14 respectively. In these figures, the force distribution and the effect of the total moment are shown.

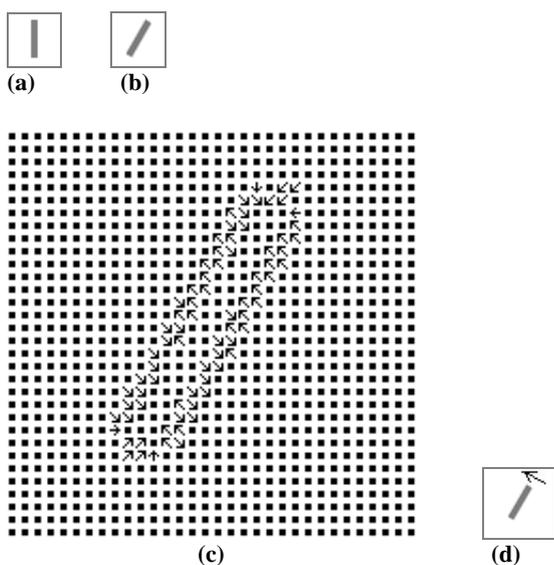

Fig. 13 The forces on the current elements in the rotated rectangle image (30 degrees), and the effect of the total moment
(a) the original rectangle image
(b) the rotated image of (a) by 30 degrees
(c) the forces on the virtual current elements in the rotated image
(d) the rotating direction caused by the total moment on the rotated image

Fig. 13(a) is the original image of the size 32×32. Fig. 13(b) shows the rotated image. Fig. 13(c) shows the force distribution in the rotated image. In Fig. 13(c), each arrow shows the discrete direction of the virtual force on the virtual current element at that position, and the black dots indicate the positions without virtual current element. If carefully observed, it can be estimated from Fig. 13(c) that the overall effect of the forces is to rotate the rectangle counterclockwise back to its original position as in Fig. 13(a). The calculating result of the total moment also proves such effect, which is shown in Fig. 13(d). The total moment for Fig. 13(b) is negative. In the screen coordinate, it has the effect of turning Fig. 13(b) counterclockwise.

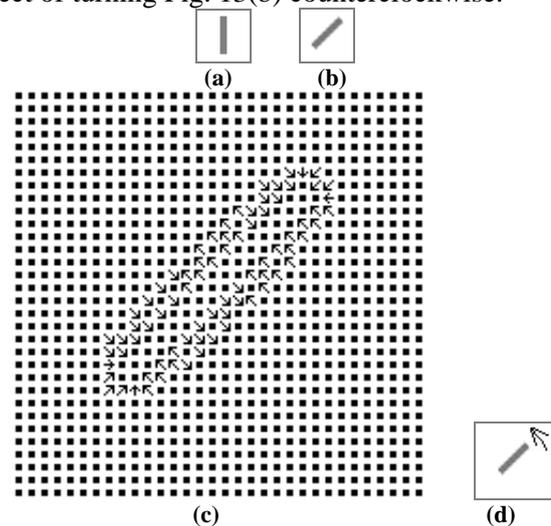

Fig. 14 The forces on the current elements in the rotated rectangle image (45 degrees), and the effect of the total moment
(a) the original rectangle image
(b) the rotated image of (a) by 45 degrees
(c) the forces on the virtual current elements in the rotated image
(d) the rotating direction caused by the total moment on the rotated image

Fig. 14 shows the result of rotating the test image by 45 degrees. Fig. 14(a) is the original image. Fig. 14(b) shows the rotated image. Fig. 14(c) shows the force distribution in the rotated image. The total moment calculated for Fig. 14(b) is negative. In the screen coordinate, it has the effect of turning Fig. 14(b) counterclockwise.

Similar experiments have also been carried out for real world images. One example of the experiments is shown in Fig. 15. Fig. 15(a) shows the broadcaster image of the size $128\times 128$. Fig. 15(b) shows the significant edge lines extracted, which represent the virtual current wires in the image. Fig. 15(c) and Fig. 15(e) show the results of rotating Fig. 15(a) by 15 degrees clockwise and counterclockwise respectively. The virtual force distribution on the rotated image is calculated, and the total moment is calculated. The total moment on Fig. 15(c) applied by the original image is positive, and in the screen coordinate it has the effect of turning Fig. 15(c) clockwise. The total moment on Fig. 15(e) is negative, and in the screen coordinate it has the effect of turning Fig. 15(e) counterclockwise.

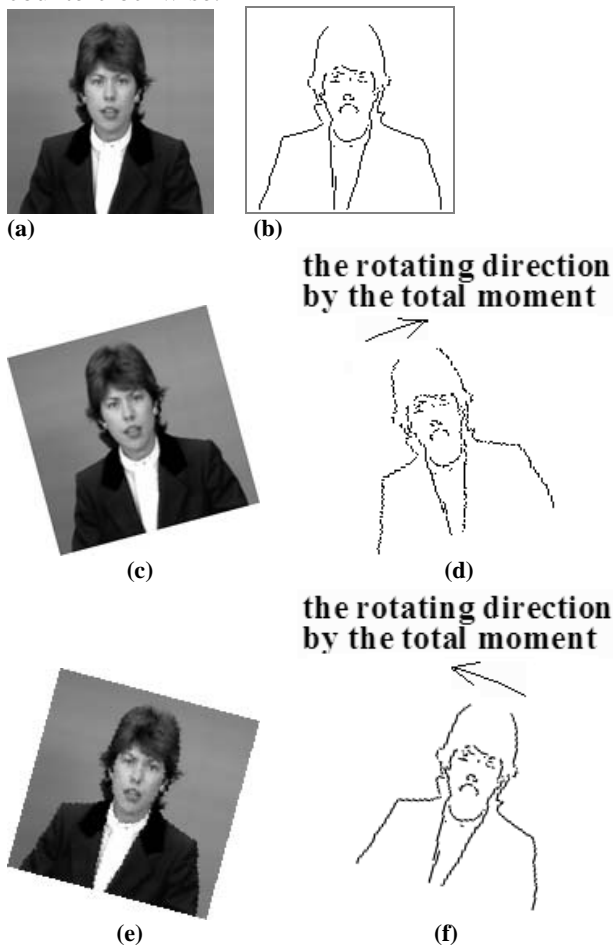

**Fig. 15** The total moments on the rotated image of the broadcaster (15 degrees clockwise and counterclockwise)
(a) the broadcaster image
(b) the significant edge lines of (a)
(c) the rotated image of (a) by 15 degrees counterclockwise
(d) the rotating direction caused by the total moment on (c)
(e) the rotated image of (a) by 15 degrees clockwise
(f) the rotating direction caused by the total moment on (e)

## 5 Image Matching for Rotating Transformation by Virtual Moment

The above preliminary experiments indicate that the original image applies moment on the rotated image, and the total moment may have the effect of turning the rotated image back to the original position. Therefore, it is interesting and meaningful to investigate the moment on the rotated image at arbitrary rotating angles. In the following, simulation experiments are carried out to study the total moment on the virtual currents of the image with rotating transformation. In these experiments, the original and the rotated images are on the same plane, and the center of rotation is the image center. In order to completely investigate the effect of the total moment on the rotated image at different rotating angles, the experiment is designed as follows. Assume the image is rotated clockwise. First, divide the range of 360 degrees into 120 equal intervals. Then the program calculates the total moment on the rotated image at each discrete rotating angle. Therefore, the direction of the total moment has two possibilities: if the moment is negative, in the screen coordinate it has the effect of turning counterclockwise; if the moment is positive, in the screen coordinate it has the effect of turning clockwise. In the experiments, the sign of the total moment at each discrete rotating angle is recorded, which represents the direction of turning effect. If the total moment has the effect of turning the rotated image back to the original position, the matching for rotating transformation can be achieved.

An example of the experiment results is shown in Fig. 16. Fig. 16(a) is the original image of a broadcaster, which has the size of $128\times 128$. Fig. 16(b) shows the significant edge line extracted, which represents the virtual currents in the image. Fig. 16(c) shows the sign distribution of the moment with respect to the rotating angle. The *x*-coordinate in Fig. 16(c) represents the number of discrete angle intervals. The *y*-coordinate represents the sign of the moment value at each rotating angle.

In Fig. 16(c), it is obvious that in this experiment the sign distribution of the total moment is somewhat irregular. Consider the change of moment sign with the angle increasing from zero. In Fig. 16(c), in a small angle range from the first interval to the seventh interval on the *x*-coordinate, the sign of moment is negative. If the angle range of 360 degrees is divided into 120 intervals, one interval corresponds to 3 degrees. Therefore, it means that in the angle range of $(0°,21°]$ the total moment is negative, which has the effect of turning the rotated image counterclockwise. In another word, if the image Fig. 16(a) is rotated clockwise in the angle range of $(0°,21°]$, the total moment on the rotated image will turn it back to the original position, which just achieves the purpose of matching. The similar case can be found in the angle range of [339°, 360°) on the *x*-coordinate. Rotating the image clockwise in the angle range of [339°, 360°) virtually corresponds to a anticlockwise rotation in the range of (0°, 21°]. In this angle range, the sign of total moment is positive, which has the effect of turning clockwise back to the original position.

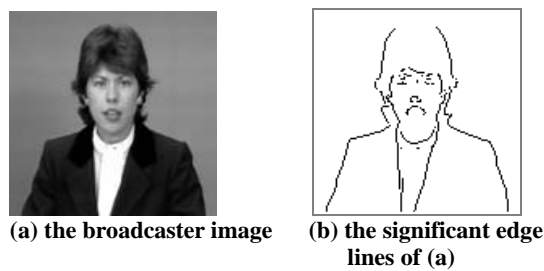

(a) the broadcaster image    (b) the significant edge lines of (a)

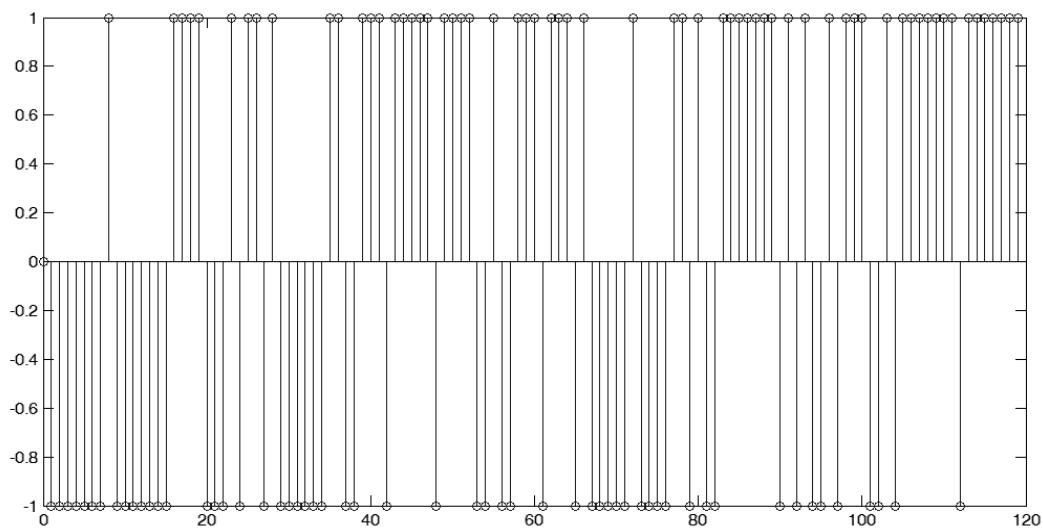

(c) the sign of total moment at various deviation angles from the original position
Fig. 16 The sign distribution of the total moment on the rotated broadcaster image with respect to the deviation angle

According to Fig. 16(c), if the image Fig. 16(a) is rotated in a limited range of angle (i.e. zero to 21 degrees clockwise or anticlockwise), the total moment has the effect of restore the rotated image back to the original position. For a rotated image in the above angle range, if it is turned continuously following the guidance of the total moment, it will finally arrive at the balance position of zero deviation angle (i.e. the original position). The original position with zero deviation angle is called a balance position because any small deviation from it will be "corrected" by the guidance of the moment.

However, in Fig. 16(c) it is obvious that for relatively large rotating angles, the total moment will not work for the matching purpose. For practical use, the above approach needs improvement. In the above experiments, the original image and the rotated one are on the same plane. In order to improve the effectiveness of the method, the authors have attempted to increase the distance between the two images in 3D space.

## 6 Improvement of the Matching Effectiveness in 3D Space

In order to improve the moment-based matching method for rotating transformation, new experiments are designed as follows. Besides the rotating transformation on the *x-y* plane, the original image and the rotated image have some distance on the *z*-coordinate. In the following experiments, the centers of the two images are both on the *z*-axis, but the images are on two different planes. In another word, the images are parallel but their heights on *z*-coordinate are different. In the experiments, only the *x* and *y* components of the virtual force are considered. The sign distributions of the total moment with different distance between the images are investigated. Some of the experiment results are shown in Fig. 17 to Fig. 20.

For the broadcaster image shown in Fig. 16(a), Fig. 17 to Fig. 20 show the results of moment sign distribution with several height distances between the original and rotated images. The height distances are 10, 20, 30 and 40 respectively. It is obvious that with the increasing height distance between the images, the moment sign distribution becomes more and more regular.

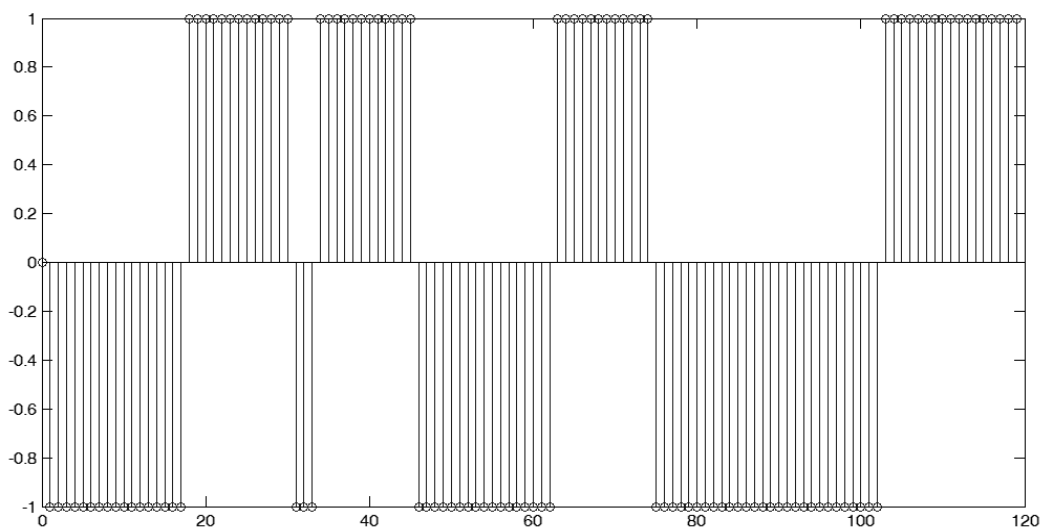

**Fig. 17** The sign distribution of the total moment on the rotated broadcaster image with respect to the deviation angle (the distance between the images is 10; the convergence range is (0°,51°] and [309°,360°). )

In a sign distribution diagram such as Fig. 17, the *x*-axis can be divided into several different sections according to the moment sign. The sign values within a section are the same, but the sign values are different in two adjacent sections. In Fig. 17, there are eight sections. There are 6, 6, and 2 sections in Fig. 18, Fig. 19, and Fig. 20 respectively. From Fig. 17 to Fig. 20, the number of sections has the decreasing tendency with the increasing distance between the images to be matched.

In each sign distribution diagram, the first and last sections along the *x*-axis are two important sections which indicate the effectiveness of the moment-based matching approach. For example, the first section in Fig. 17 is (0,17] on the *x*-axis, which corresponds to an angle range of (0°,51°] (here an interval on the *x*-axis corresponds to 3 degrees). The last section in Fig. 17 corresponds to an angle range of [309°,360°]. Similar to the analysis of Fig. 16(c), the deviation angles within the first section correspond to clockwise rotation from the original position, and those within the last section correspond to anticlockwise rotation from the original position. If the rotation angle of image is within these two sections, the moment will have the effect of turning the rotated image back to the original position (with a zero rotation angle), which has been discussed in the analysis of Fig. 16(c). But if the rotation angle is not within the above two sections, the moment-based method does not work for matching. In another word, if the rotation angle exceeds the range covered by the above two sections, the rotated image can not be turned back to the original position under the guidance of the moment sign.

Therefore, the first and last sections in a moment sign distribution diagram are defined as the convergence range of rotating angles. For an image, if the convergence range gets larger, the moment-based method becomes more effective. For the experiment results of the broadcaster image, it can be found in Fig. 16(c) and Fig. 17 to Fig. 20 that the convergence range has an increasing tendency with the increasing distance between the images. Therefore, the increasing of height distance between images improves the method's effectiveness. When the height distance is increased to 40, the convergence range covers the whole *x*-axis of 360°, which means that the moment-based method works for any rotating angle in the matching.

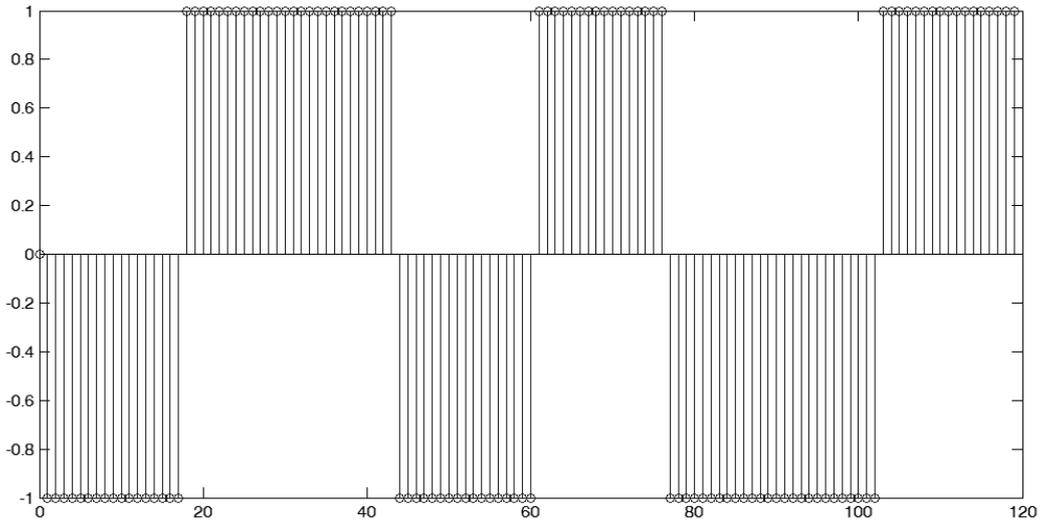

**Fig. 18** The sign distribution of the total moment on the rotated broadcaster image with respect to the deviation angle (the distance between the images is 20)

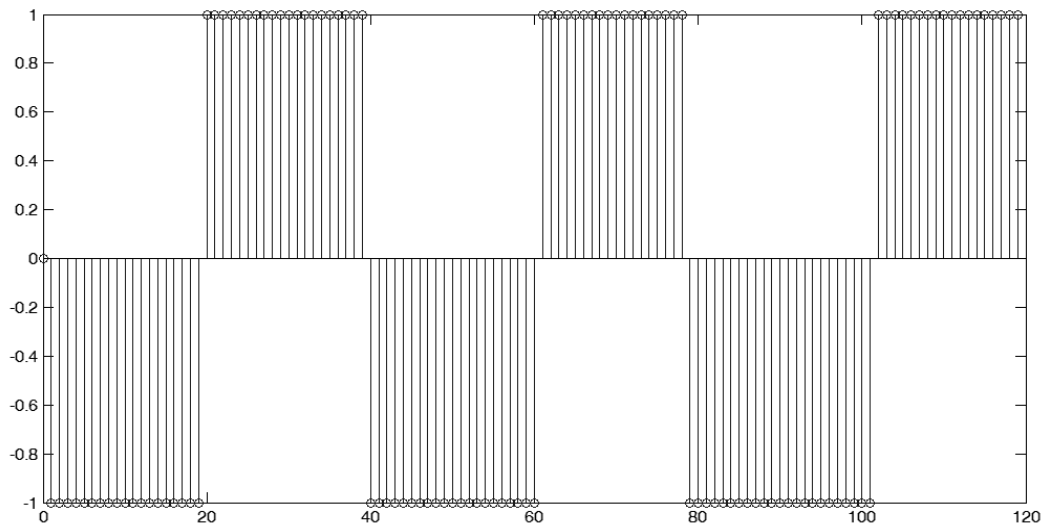

**Fig. 19** The sign distribution of the total moment on the rotated broadcaster image with respect to the deviation angle (the distance between the images is 30; the convergence range is (0°,57°] and [303°,360°). )

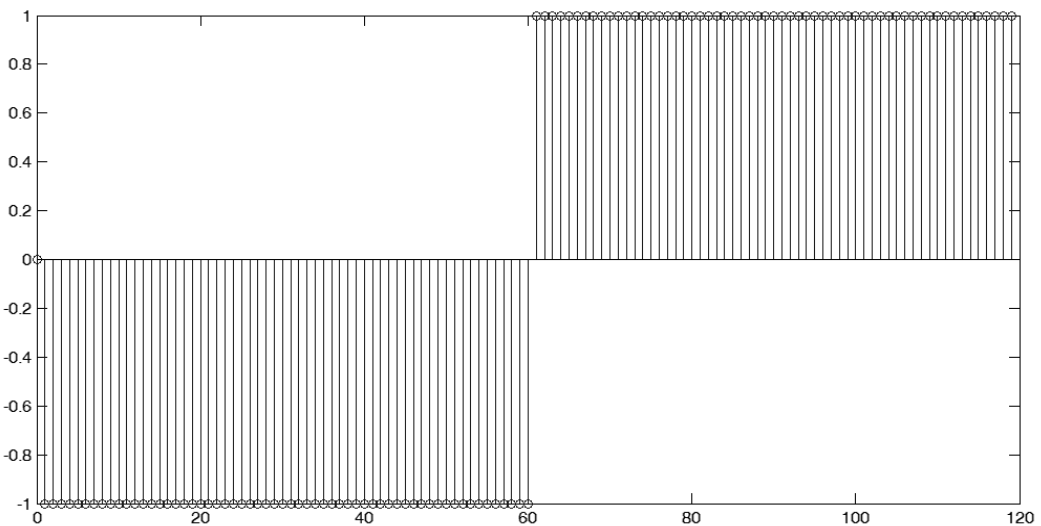

**Fig. 20** The sign distribution of the total moment on the rotated broadcaster image with respect to the deviation angle (the distance between the images is 40; the convergence range is (0°, 360°). )

In order to show the moment sign distribution more intuitively, the sign distribution diagram such as Fig. 20 can be displayed in a way like the pie chart. Fig. 21 shows the pie chart form of Fig. 20. In Fig. 21, the whole circle (representing the whole range of 360 degrees on the *x*-axis in the sign distribution diagram) is divided into sections, which just correspond to the sections on the *x*-axis in the sign distribution diagram. Each section on the circle in Fig. 21 is represented by a grayscale different from the adjacent ones, and each section corresponds to an angle range. The arrow on each section shows the direction of the moment's rotating effect if the image is rotated at an angle within that section. The "valid" sections in the pie chart form correspond to the convergence range.

According to Fig. 21, if the broadcaster image is rotated clockwise (in the (0°, 180°] range on the *x*-axis), the effect of moment will turn it counterclockwise until arriving at the original position with zero rotating angle. On the other hand, if the broadcaster image is rotated counterclockwise (in the (180°, 360°) range on the *x*-axis), the effect of moment will turn it clockwise until arriving at the original position. It is indicated by Fig. 21 that, for the broadcaster image, when the height distance between the two images is 40, the moment can turn the rotated image back to the original position whether it is rotated clockwise or counterclockwise. Some of the experiment results on other real world images of the size $128 \times 128$ (the images of cameraman, peppers, locomotive, boat, and a medical image of brain) are shown as follows.

Fig. 22(a) shows the original cameraman image. Fig. 22(b) shows the sign distribution of total moment with the height distance 10, and the corresponding pie chart form is shown in Fig. 22(c). It can be found in Fig. 22(b) and Fig.22(c) that the convergence range covers all the 360 degrees just like the result in Fig. 21.

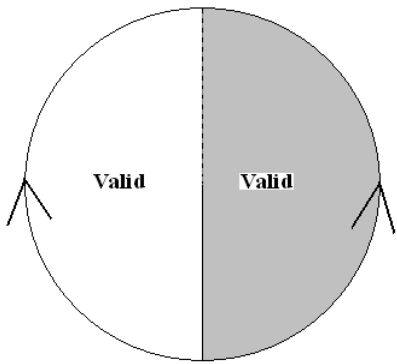

**Fig. 21 The rotation direction according to the total moment under different deviation angles (the distance between the images is 40)**

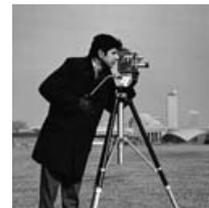

**Fig. 22(a) The cameraman image**

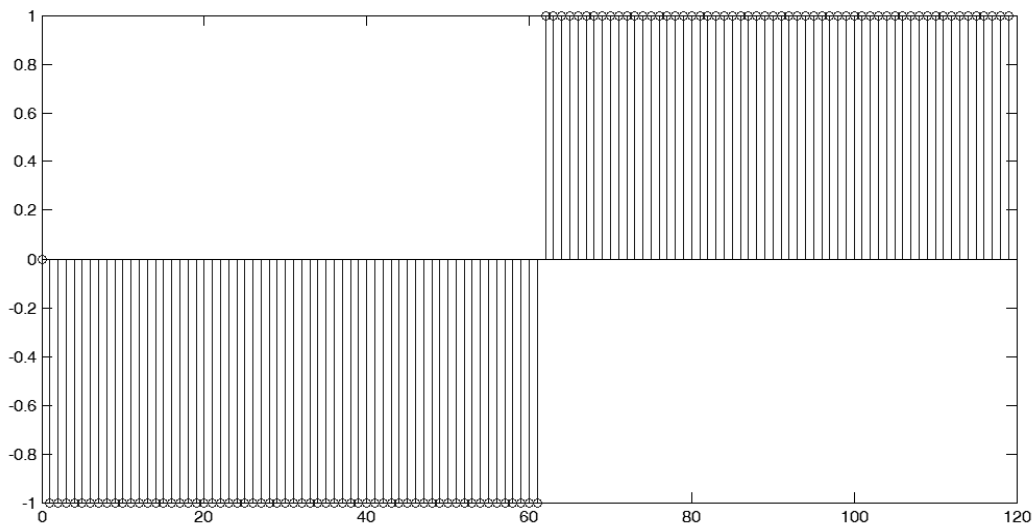

**Fig. 22(b) The sign distribution of the total moment on the rotated cameraman image with respect to the deviation angle (the distance between the images is 10; the convergence range is (0°, 360°). )**

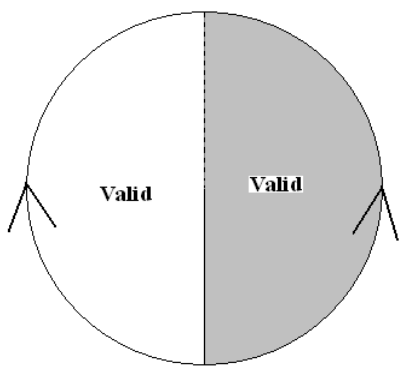

**Fig. 22(c) The rotation direction according to the sign of moment under different deviation angles (the distance between the images is 10)**

Fig. 23(a) shows the original peppers image. Fig. 23(b) shows the sign distribution of total moment with the height distance 20, and the corresponding pie chart form is shown in Fig. 23(c). It can be found in Fig. 23(b) and Fig. 23(c) that the convergence range covers all the 360 degrees. But the first and last sections in the convergence range of Fig. 23(b) is somewhat different from Fig. 20 and Fig. 22(b), because the dividing point between the two sections in Fig. 23(b) is not 180 degrees (i.e. the middle of the whole 360° range). However, it does not affect the validity of the moment-based method. Whatever the rotating angle is, the moment can finally guide the rotated image back to the original position of zero deviation angle.

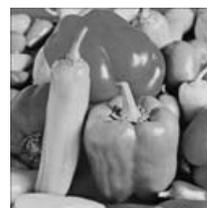

**Fig. 23(a) The peppers image**

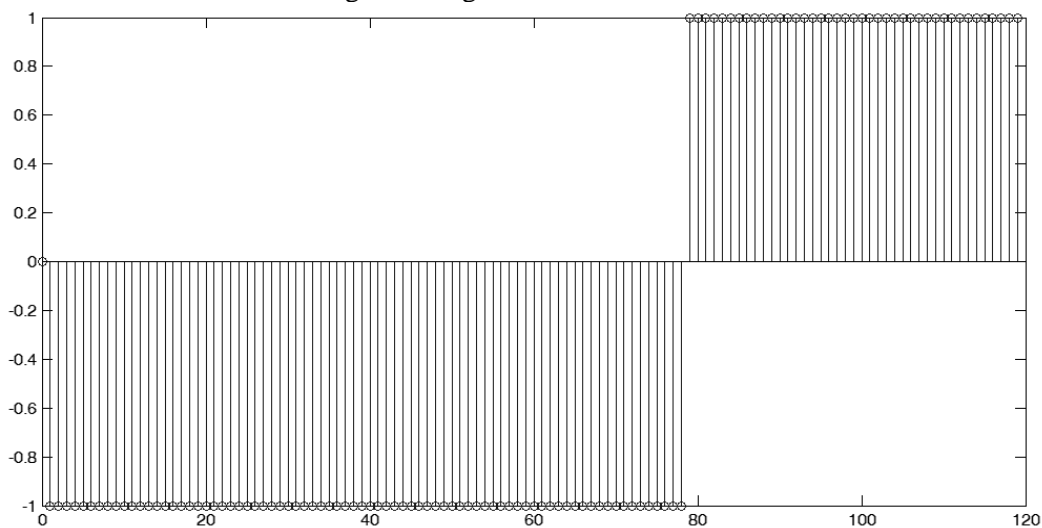

**Fig. 23(b) The sign distribution of the total moment on the rotated peppers image with respect to the deviation angle (the distance between the images is 20; the convergence range is (0°, 360°).)**

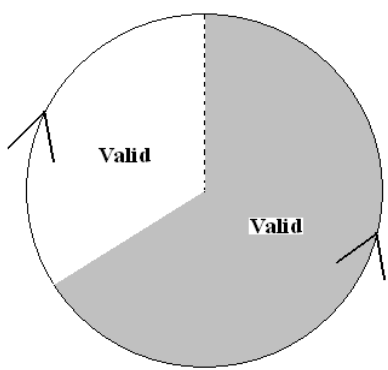

**Fig. 23(c) The rotation direction according to the sign of moment under different deviation angles (the distance between the images is 20)**

The experiment result for the locomotive image is similar to that of the peppers image. Fig. 24(a) shows the original locomotive image. Fig. 24(b) shows the sign distribution of total moment with the height distance 30 between the images, and the corresponding pie chart form is shown in Fig. 24(c). It can be found in Fig. 24(b) and Fig. 24(c) that the convergence range covers all the 360 degrees. But just like Fig. 23(b), the dividing point between the two sections in Fig. 24(b) is not 180 degrees.

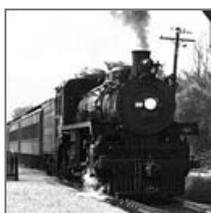

**Fig. 24(a) The locomotive image**

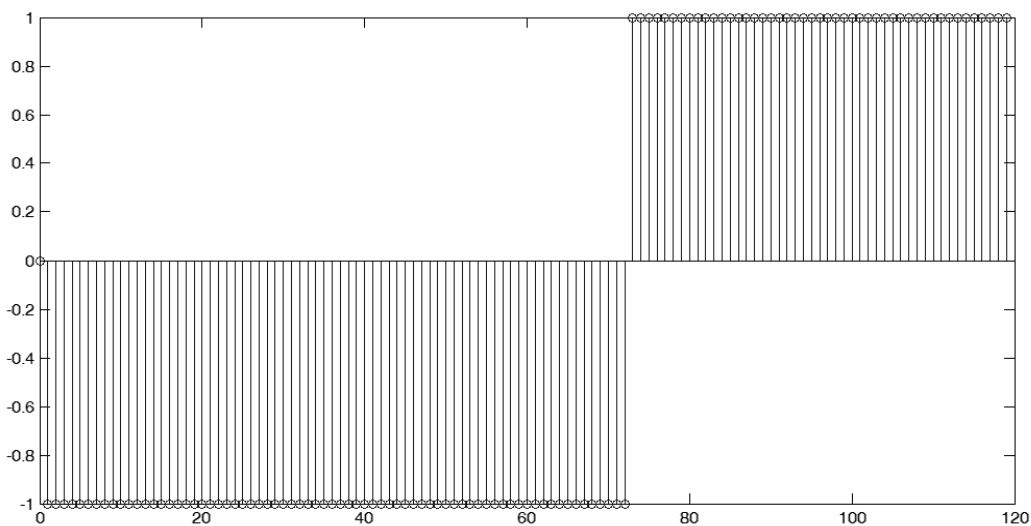

**Fig. 24(b) The sign distribution of the total moment on the rotated locomotive image with respect to the deviation angle (the distance between the images is 30; the convergence range is (0°, 360°). )**

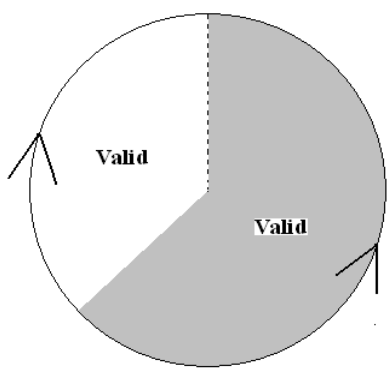

**Fig. 24(c) The rotation direction according to the sign of moment under different deviation angles (the distance between the images is 30)**

The experimental results for the brain image and boat image are quite different from the above ones. Fig. 25(a) shows the original medical image of the brain. Fig. 25(b) shows the sign distribution of total moment with the height distance 30, and the corresponding pie chart form is shown in Fig. 25(c). In Fig. 25(b) there are four sections, and the convergence range (the first and last section) does not cover the whole 360° range. Fig. 25(c) gives a clear view. For a deviation angle in the "invalid" sections in Fig. 25(c), the moment-based method is invalid because the effect of moment will turn the rotated image to a locally balanced point shown as the "oscillating angle" in Fig. 25(c). It is called "locally balanced" because when it is reached the image will be turned back and forth around it according to the effect of the moment, which can be clearly seen in Fig. 25(c). Therefore, for the medical image of brain here, the effectiveness of the moment-based method is limited (within the convergence range labeled as "valid" in Fig. 25(c)).

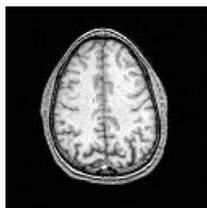

**Fig. 25(a) The medical image of the brain**

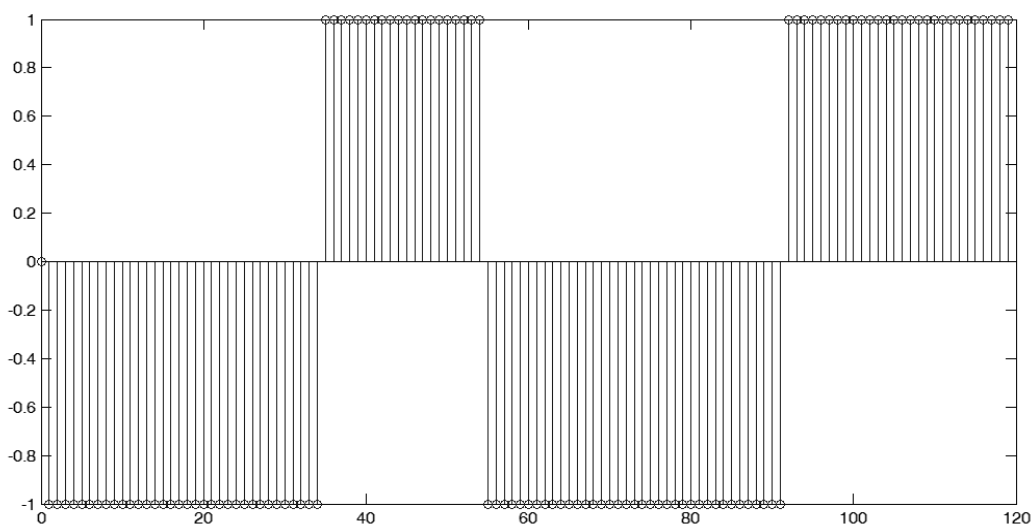

**Fig. 25(b) The sign distribution of the total moment on the rotated brain image with respect to the deviation angle (the distance between the images is 30; the convergence range is (0°,102°] and [258°,360°).)**

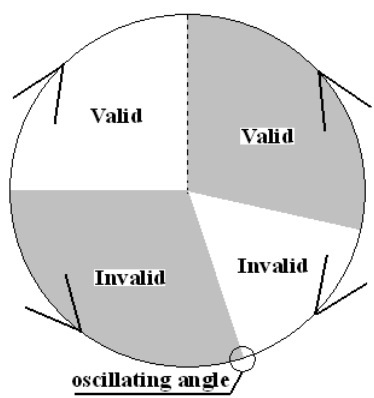

**Fig. 25(c) The rotation direction according to the sign of moment under different deviation angles (the distance between the images is 30)**

The experiment result for the boat image is similar to that of the brain image. Fig. 26(a) shows the original boat image. Fig. 26(b) shows the sign distribution of total moment with the height distance 20, and the corresponding pie chart form is shown in Fig. 26(c). In Fig. 26(b) there are four sections, and the convergence range does not cover the whole 360° range. Fig. 26(c) gives a clear view.

In the "invalid" sections in Fig. 26(c), the moment-based method is invalid because the effect of moment will turn the rotated image to a locally balanced point shown as the "oscillating angle" in Fig. 26(c), just like the case in Fig. 25(c). Therefore, for the boat here, the effectiveness of the moment-based method is also limited (within the convergence range labeled as "valid" in Fig. 26(c)).

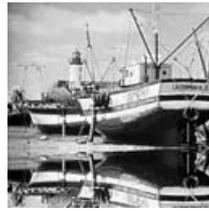

**Fig. 26(a) The boat image**

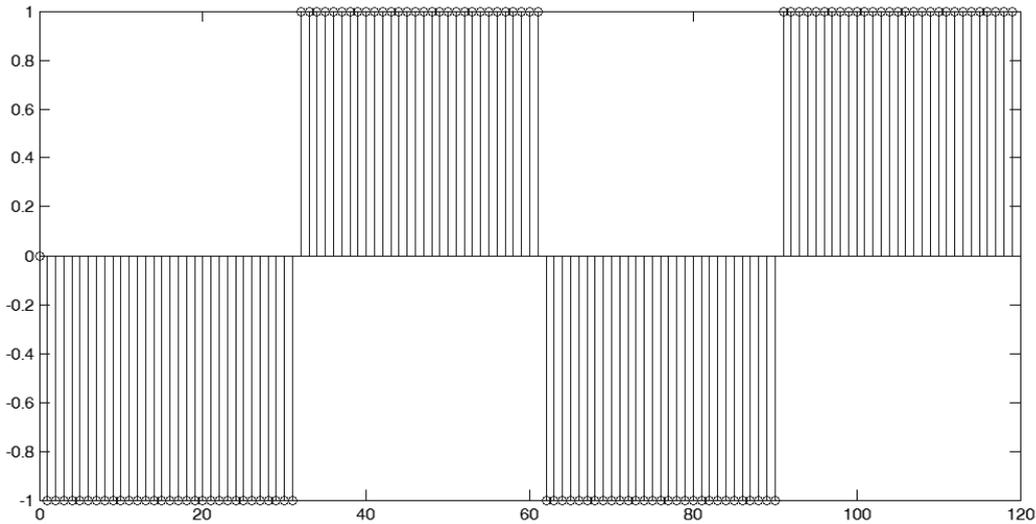

**Fig. 26(b) The sign distribution of the total moment on the rotated boat image with respect to the deviation angle (the distance between the images is 20; the convergence range is (0°,93°] and [267°,360°). )**

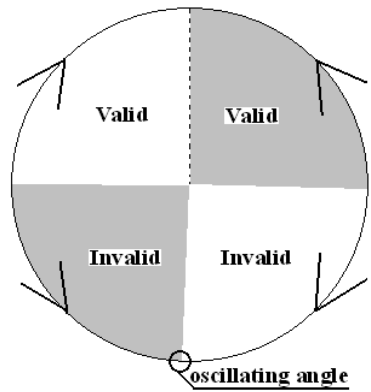

**Fig. 26(c) The rotation direction according to the sign of moment under different deviation angles (the distance between the images is 20)**

The above experiments show that the moment-based method can work as a novel matching approach for rotating transformation within the convergence range of deviation angle. According to the above experiments, the movement-based matching method is summarized as following:

***Step*1**: Extract the significant edge lines as the virtual currents in the original and rotated images.
***Step*2**: Calculate the force on each current element in the rotated image applied by the original image.
***Step*3**: Calculate the total moment based on the force on each current element in the rotated image.
***Step*4**: Turn the rotated image by a predefined small angle around the image center according to the total moment. If a balanced point is reached, the process ends; otherwise, return to ***Step*1**. Reaching a balanced point means that the turning process becomes oscillating back and forth around a certain angle, such as the case at the zero deviation angle.

If the rotating angle is within the convergence range, the proposed method can effectively achieve image matching for rotating transformation.

# 7 Conclusion and Discussion

In physics, current is a basic source of magnetic field. Moreover, the magnetic field applies force on other currents in it. The interesting phenomena of electromagnetic interaction can generate inspiring physical effects, which may be exploited in design novel image processing algorithms. A macroscopic current in the wire is based on the microscopic movement of electrons. The virtual current proposed in the paper is a kind of reasonable imitation of physical current in the image space. In another word, the virtual edge current in images is a discrete simulation of the physical current in the discrete image space, which is a discrete flow field running along the edge lines (i.e. the isolines or level curves of grayscale) in the image.

In this paper, the significant edge lines are extracted from the edge current vectors by a Canny-like operation as the image's structure representation, based on which the virtual edge current is defined. The virtual interaction between the significant edge currents is studied by imitating the electro-magnetic interaction between the current-carrying wires. The virtual interaction between edge currents is then used in image matching for rotating transformation. The preliminary experimental results indicate the effectiveness and promising application of the proposed method for image matching.

It should be noted that the computation efficiency of the proposed method is high, which is preferable for practical tasks. Moreover, its effectiveness for large rotation angles is distinctive. However, the method proposed is for matching rotation transformation, and it should be integrated with other methods to suit the tasks in which not only rotation but also shifting transformation occur. Future work will also consider more practical issues to improve the performance of image matching with rotating transformation, such as increasing the convergence range of rotation angle in matching some real-world images.

*References:*
[1] Mark S. Nixon, Xin U. Liu, Cem Direkoglu, David J. Hurley, On using physical analogies for feature and shape extraction in computer vision, Computer Journal, Vol. 54, No. 1, 2011, pp. 11-25.
[2] D.J. Hurley, M.S. Nixon, J.N. Carter, A new force field transform for ear and face recognition, IEEE International Conference on Image Processing, Vol. 1, 2000, pp. 25-28.
[3] David J. Hurley, Mark S. Nixon, John N. Carter, Force field feature extraction for ear biometrics, Computer Vision and Image Understanding, Vol. 98, No. 3, 2005, pp. 491-512.
[4] David J. Hurley, Mark S. Nixon, John N. Carter, Force field energy functionals for image feature extraction, Image and Vision Computing, Vol. 20, No. 5-6, 2002, pp. 311-317
[5] Xin U Liu, Mark S Nixon, Water Flow Based Complex Feature Extraction. Advanced Concepts for Intelligent Vision Systems, Lecture Notes in Computer Science, 2006. pp. 833-845.
[6] Xin U Liu, Mark S Nixon, Medical Image Segmentation by Water Flow, in Proceedings of Medical Image Understanding and Analysis, MIUA 2007.
[7] Xin U Liu, Mark S Nixon, Water flow based vessel detection in retinal images, Proceedings of IET International Conference on Visual Information Engineering 2006, 2006, pp. 345-350.
[8] Xin U Liu, Mark S Nixon, Image and volume segmentation by water flow, Third International Symposium on Proceedings of Advances in Visual Computing, ISVC 2007, 2007, pp. 62-74.
[9] Lisa Gottesfeld Brown, A survey of image registration techniques, ACM Computing Surveys (CSUR) archive, Volume 24 , Issue 4, 1992, pp. 325 - 376
[10] A. Ardeshir Goshtasby, 2-D and 3-D Image Registration for Medical, Remote Sensing, and Industrial Applications, Wiley Press, 2005.
[11] Richard Szeliski, Image Alignment and Stitching: A Tutorial, Foundations and Trends in Computer Graphics and Computer Vision, 2:1-104, 2006.
[12] B. Fischer, J. Modersitzki, Ill-posed medicine - an introduction to image registration, Inverse Problems, 24:1-19, 2008
[13] Barbara Zitova, Jan Flusser: Image registration methods: a survey. Image Vision Comput. 21(11): 977-1000 (2003).
[14] A. Ardeshir Goshtasby, Image Registration: Principles, Tools and Methods, Springer, 2012.
[15] Joseph V. Hajnal, Medical Image Registration, Taylor & Francis Group, 2001.
[16] Jacqueline Le Moigne, Nathan S. Netanyahu, Roger D. Eastman, Image Registration for Remote Sensing, Cambridge University Press, 2011.
[17] Thomas B. Moeslund, Introduction to Video and Image Processing: Building Real Systems and Applications, Springer, 2012.
[18] P. Hammond, Electromagnetism for Engineers: An Introductory Course, Oxford University Press, USA, forth edition, 1997.
[19] I. S. Grant and W. R. Phillips,

Electromagnetism, John Wiley & Sons, second edition, 1990.

[20] Terence W. Barrett, Topological foundations of electromagnetism, World Scientific series in contemporary chemical physics, Vol. 26, World Scientific, 2008.

[21] Minoru Fujimoto, Physics of classical electromagnetism, Springer, 2007.

[22] J. Canny, A Computational Approach To Edge Detection, IEEE Trans. Pattern Analysis and Machine Intelligence, 8(6):679–698, 1986.

[23] R. Deriche, Using Canny's criteria to derive a recursively implemented optimal edge detector, Int. J. Computer Vision, Vol. 1, pp. 167–187, 1987.

[24] Tom W.B. Kibble, Frank H. Berkshire, Classical Mechanics (5th ed.), Imperial College Press, 2004.

[25] Daniel Kleppner, Robert J. Kolenkow, An Introduction to Mechanics, Cambridge University Press, 2010.